\documentclass[11pt,a4paper]{article}
\usepackage[hyperref]{style/acl2019}
\usepackage{times}
\usepackage{latexsym}
\usepackage{amsthm}

\aclfinalcopy % Uncomment this line for the final submission
\def\aclpaperid{757} %  Enter the acl Paper ID here

\newif\ifcomment\commenttrue
\usepackage{framed}
\usepackage{mdwlist}
\usepackage{latexsym}
\usepackage{xcolor}
\usepackage{nicefrac}
\usepackage{booktabs}
\usepackage{amsfonts}
\usepackage[T1]{fontenc}
\usepackage{bold-extra}
\usepackage{amsmath}
\usepackage{dsfont}
\usepackage{amssymb}
\usepackage{bm}
\usepackage{graphicx}
\usepackage{mathtools}
\usepackage{microtype}
\usepackage{multirow}
\usepackage{multicol}
\usepackage{latexsym,comment}

\newcommand{\gem}[1]{\mbox{\textsc{gem}}}
\newcommand{\abr}[1]{\textsc{#1}}

\newcommand{\hidetext}[1]{}
\newcommand{\ignore}[1]{}

\ifcomment
\newcommand{\pinaforecomment}[3]{\colorbox{#1}{\parbox{.8\linewidth}{#2: #3}}}
\else
\newcommand{\pinaforecomment}[3]{}
\fi

\newcommand{\smallurl}[1]{ \begin{tiny}\url{#1}\end{tiny}}

\definecolor{lightblue}{HTML}{3cc7ea}
\definecolor{CUgold}{HTML}{CFB87C}
\definecolor{grey}{rgb}{0.95,0.95,0.95}
\definecolor{ceil}{rgb}{0.57, 0.63, 0.81}

\ifcomment
\newcommand{\mzcomment}[1]{ \colorbox{lightblue}{   \parbox{.8\linewidth}{ MZ: #1}  }}
\else
\newcommand{\mzcomment}[1]{}
\fi

\newcommand{\vect}[1]{\bm{\mathbf{#1}}}

\newcommand{\figfile}[1]{figures/#1}

\newtheorem{theorem}{Theorem}
\newtheorem{assumption}{Assumption}

\usepackage{CJKutf8}
\newcommand{\ja}[1]{\begin{CJK*}{UTF8}{ipxm}#1\end{CJK*}}

\newcommand{\name}[0]{Iterative Normalization}

\usepackage{array}
\newcolumntype{R}[0]{>{\raggedleft\let\newline\\\arraybackslash\hspace{0pt}}p{1cm}}

\newcommand{\flag}[1]{{\setlength{\fboxsep}{0pt}\fbox{\includegraphics[height=0.24cm,width=0.36cm]{\figfile{flags/#1.pdf}}}}}

\title{Are Girls Neko or Sh\={o}jo? Cross-Lingual Alignment of Non-Isomorphic Embeddings with Iterative Normalization}

\author{
  \begin{tabular}{ccc}
    Mozhi Zhang$^1$ & Keyulu Xu$^2$ & Ken-ichi Kawarabayashi$^3$
  \end{tabular}\\
  \begin{tabular}{cc}
    \textbf{Stefanie Jegelka$^2$} & \textbf{Jordan Boyd-Graber$^1$}
  \end{tabular}\\
  $^1$University of Maryland, College Park, Maryland, USA\\
  $^2$Massachusetts Institute of Technology, Cambridge, Massachusetts, USA\\
  $^3$National Institue of Informatics, Tokyo, Japan\\
  \begin{tabular}{ccc}
    {\small\tt \{mozhi,jbg\}@umiacs.umd.edu} & {\small\tt \{keyulu,stefje\}@mit.edu} & {\small\tt k\_keniti@nii.ac.jp}
  \end{tabular}
}

\begin{document}

\maketitle
% abstract

\begin{abstract}

  Cross-lingual word embeddings~(\abr{clwe}) underlie many
  multilingual natural language processing systems, often through
  orthogonal transformations of pre-trained monolingual embeddings.
  However, orthogonal mapping only works on language pairs
  whose embeddings are naturally isomorphic.
  For non-isomorphic pairs, our method (\name{}) transforms
  monolingual embeddings to make orthogonal alignment easier by
  simultaneously enforcing that (1)~individual word vectors are unit
  length, and (2)~each language's average vector is zero.
  \name{} consistently improves word translation accuracy of three
  \abr{clwe} methods, with the largest improvement observed on English-Japanese
  (from 2\% to 44\% test accuracy).

\end{abstract}

% intro

\section{Orthogonal Cross-Lingual Mappings}\label{sec:intro}

Cross-lingual word embedding (\abr{clwe}) models map words from
multiple languages to a shared vector space, where words with similar
meanings are close, regardless of language.
\abr{clwe} is widely used in multilingual natural language
processing~\citep{klementiev-12,guo-15,zhang-16}.
Recent \abr{clwe} methods~\citep{ruder-17,glavas-19} independently train two
monolingual embeddings on large monolingual corpora and then align them
with a linear transformation.
Previous work argues that these transformations should be
\emph{orthogonal}~\citep{xing-15,smith-17,artetxe-16}: for any two words, the
dot product of their representations is the same as the dot product with the
transformation.
This preserves similarities and substructure of the original monolingual word
embedding but enriches the embeddings with multilingual connections between
languages.

\begin{figure*}[!t]
  \centering
  \includegraphics[width=.8\linewidth]{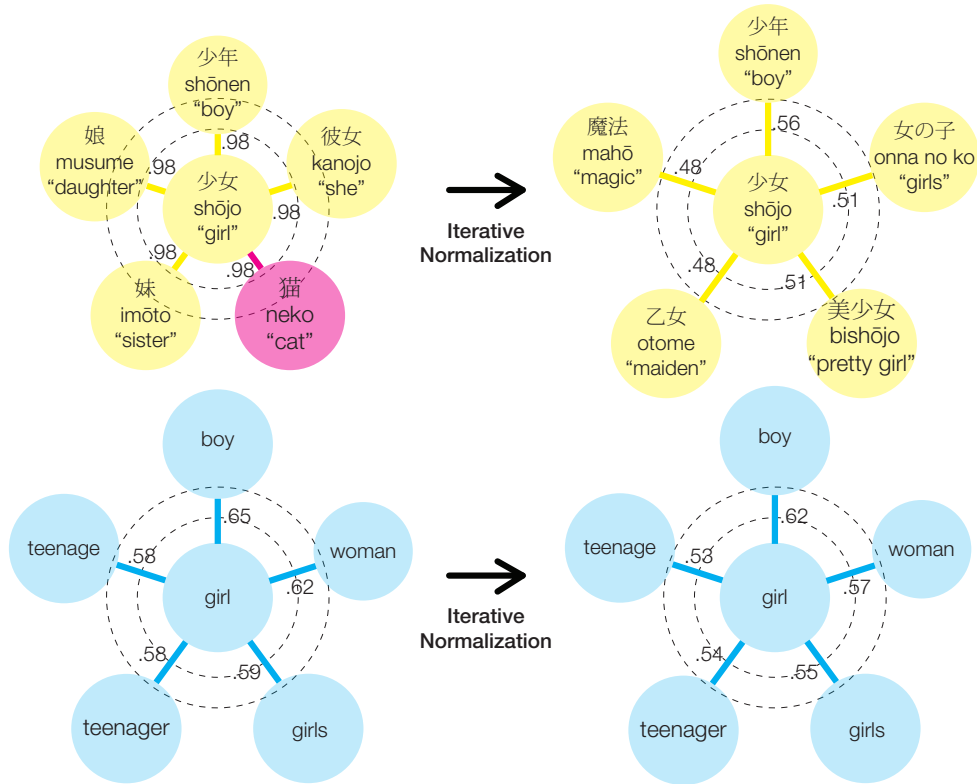}
  \caption{The most similar Japanese words for \ja{少女} (sh\={o}jo ``girl'')
  and English words for ``girl'', measured by cosine similarity on Wikipedia
  fastText vectors, before (left) and after (right) \name{}.
  In the original embedding spaces, ``boy'' is the nearest neighbor for both
  languages but with a very different cosine similarity, and ``cat'' in English
  is not close to ``girl'': both violate the isomorphism assumed by an
  orthogonal transformation for cross-lingual representations.
  \name{} replaces \ja{猫} (neko ``cat'') with the more
  relevant \ja{美少女} (bish\={o}jo ``pretty girl'') and brings cosine similarities 
  closer.}
  \label{fig:example}
\end{figure*}

Thus, many state-of-the-art mapping-based \abr{clwe} methods impose an
orthogonal
constraint~\citep{artetxe-17,conneau-18,alvarez-18,artetxe-18b,ruder-18,alvarez-19}.
The success of orthogonal methods relies on the assumption that embedding spaces
are isomorphic; i.e., they have the same inner-product structures across
languages, but this does not hold for all
languages~\citep{sogaard-18,fujinuma-19}.
For example, English and Japanese fastText vectors~\citep{bojanowski-17} have
different substructures around ``girl'' (Figure~\ref{fig:example} left).
As a result, orthogonal mapping fails on some languages---when
\citet{hoshen-18} align fastText embeddings with orthogonal mappings, they
report 81\% English--Spanish word translation accuracy but only 2\% for the
more distant English--Japanese.

While recent work challenges the orthogonal
assumption~\citep{doval-18,joulin-18,jawanpuria-19}, we focus on whether simple
preprocessing techniques can \emph{improve the suitability of orthogonal
models}.
Our iterative method normalizes monolingual embeddings to make their structures
more similar (Figure~\ref{fig:example}), which improves subsequent
alignment.

Our method is motivated by two desired properties of monolingual embeddings
that support orthogonal alignment:
\begin{enumerate*}
\item Every word vector has the same length.
\item Each language's mean has the same length.
\end{enumerate*}
Standard preprocessing such as dimension-wise mean centering and length
normalization~\cite{artetxe-16} do not meet the two requirements at
the same time.
Our analysis leads to \emph{\name{}}, an alternating projection
algorithm that normalizes any word embedding to provably satisfy both
conditions.\footnote{The code is available at \url{https://github.com/zhangmozhi/iternorm}.}
After normalizing the monolingual embeddings, we then apply
mapping-based \abr{clwe} algorithms on the transformed embeddings.

We empirically validate our theory by combining \name{} with three
mapping-based \abr{clwe} methods.
\name{} improves word translation accuracy on a dictionary induction benchmark
across thirty-nine language pairs.

\section{Learning Orthogonal Mappings}\label{sec:prelim}

This section reviews learning orthogonal
cross-lingual mapping between word embeddings and, along the way,
introduces our
notation.

We start with pre-trained word embeddings in a source language and a
target language.
We assume\footnote{Word translation benchmarks
use the same assumptions.}
 all embeddings are~$d$-dimensional, and the two languages
have the same vocabulary size~$n$.
Let $\vect{X} \in \mathbb{R}^{d \times n}$ be the word embedding
matrix for the source language, where each column $\vect{x}_i \in
\mathbb{R}^d$ is the representation of the $i$-th word from the source
language, and let $\vect{Z} \in \mathbb{R}^{d \times n}$ be the word
embedding matrix for the target language.
Our goal is to learn a transformation matrix~$\vect{W} \in
\mathbb{R}^{d\times d}$ that maps the source language vectors to the
target language space.
While our experiments focus on the supervised case with a seed dictionary $\mathcal{D}$
with translation pairs $(i, j)$, the analysis also applies to
unsupervised projection.

One straightforward way to learn $\vect{W}$ is by minimizing Euclidean
distances between translation pairs~\citep{mikolov-13b}.  Formally, we
solve:
\begin{equation}\label{eq:obj}
\min_{\vect{W}} \sum_{(i,j) \in \mathcal{D}} \| \vect{W} \vect{x}_i - \vect{z}_j \|_2^2.
\end{equation}

\citet{xing-15} further restrict $\vect{W}$ to orthogonal
transformations; i.e., $\vect{W}^\top \vect{W} = \vect{I}$.
The orthogonal constraint significantly improves word translation
accuracy~\citep{artetxe-16}.
However, this method still fails for some
language pairs because word embeddings are not isomorphic across
languages.
To improve orthogonal alignment between non-isomorphic embedding
spaces, we aim to transform monolingual embeddings in a way that helps
orthogonal transformation.

\section{When Orthogonal Mappings Work}
\label{sec:property}

When are two embedding spaces easily aligned? A good
orthogonal mapping is more likely if word vectors have two
properties: \emph{length-invariance} and \emph{center-invariance}.

\paragraph{Length-Invariance.}
First, all word vectors should have the same, constant
length.
Length-invariance resolves inconsistencies between monolingual word
embedding and cross-lingual mapping objectives~\citep{xing-15}.
During training, popular word embedding
algorithms~\cite{mikolov-13-fixed,pennington-14,bojanowski-17} maximize
\emph{dot products} between similar words, but evaluate on
\emph{cosine similarity}.
To make things worse, the transformation matrix 
minimizes a third metric, \emph{Euclidean
  distance}~(Equation~\ref{eq:obj}).
This inconsistency is naturally resolved when the lengths of word
vectors are fixed.
Suppose $\vect{u} \in \mathbb{R}^d$ and $\vect{v} \in \mathbb{R}^d$
have the same length, then
\begin{equation*}
\vect{u}^\top \vect{v} \propto \cos(\vect{u},\vect{v})
= 1 - \frac{1}{2} \| \vect{u} - \vect{v} \|_2^2.
\end{equation*}
Minimizing Euclidean distance is equivalent to maximizing both dot
product and cosine similarity with constant word vector lengths, thus
making objectives consistent.

Length-invariance also satisfies a prerequisite for bilingual
orthogonal alignment: the embeddings of translation pairs should have the same
length.
If a source word vector $\vect{x}_i$ can be aligned to its target language
translation $\vect{z}_j = \vect{W} \vect{x}_i$ with an orthogonal matrix
$\vect{W}$, then 
\begin{equation}\label{eq:len}
  \| \vect{z}_j \|_2 = \| \vect{W} \vect{x}_i \|_2 = \| \vect{x}_i \|_2,
\end{equation}
where the second equality follows from the orthogonality of $\vect{W}$.
Equation~\eqref{eq:len} is trivially satisfied if all vectors have the same
length.
In summary, length-invariance not only promotes consistency between
monolingual word embedding and cross-lingual mapping objective but
also simplifies translation pair alignment.

\paragraph{Center-Invariance.}
Our second condition is that the mean vector of different languages should have
the same length, which we prove is a pre-requisite for orthogonal
alignment.
Suppose two embedding matrices $\vect{X}$ and $\vect{Z}$ can be aligned with an
orthogonal matrix $\vect{W}$ such that $\vect{Z} = \vect{W} \vect{X}$.
Let $\vect{\bar{x}} = \frac{1}{n} \sum_{i=1}^n \vect{x}_i$ and $\vect{\bar{z}}
= \frac{1}{n} \sum_{i=1}^n \vect{z}_i$ be the mean vectors.
Then $\vect{\bar{z}} = \vect{W} \vect{\bar{x}}$.  Since $\vect{W}$ is 
orthogonal, 
\begin{equation*}
  \| \vect{\bar{z}} \|_2 = \| \vect{W} \vect{\bar{x}} \|_2 = \| \vect{\bar{x}}\|_2.
\end{equation*}
In other words, orthogonal mappings can \emph{only} align embedding spaces
with equal-magnitude centers.

A stronger version of center-invariance is zero-mean, where the mean vector of
each language is zero.
\newcite{artetxe-16} find that centering improves dictionary induction; our
analysis provides an explanation.

\begin{table*}
\centering
\begin{tabular}{llRRRRRRR}
\toprule
Method & Normalization & \flag{ja}~\abr{ja} & \flag{zh}~\abr{zh} & \flag{hi}~\abr{hi} & \flag{tr}~\abr{tr} & \flag{da}~\abr{da} & \flag{de}~\abr{de} & \flag{es}~\abr{es}\\
\midrule
Procrustes & None & 1.7 & 32.5 & 33.3 & 44.9 & 54.0 & 73.5 & 81.4\\
& \abr{c+l} & 12.3 & 41.1 & 34.0 & 46.5 & 54.9 & 74.6 & 81.3\\
& \abr{in} & \textbf{44.3} & \textbf{44.2} & \textbf{36.7} & \textbf{48.7} & \textbf{58.4} & \textbf{75.5} & \textbf{81.5}\\
\midrule
Procrustes + refine & None & 1.7 & 32.5 & 33.6 & 46.3 & 56.8 & 74.3 & 81.9\\
& \abr{c+l} & 13.1 & 42.3 & 34.9 & 48.7 & 59.3 & 75.2 & 82.4\\
& \abr{in} & \textbf{44.3} & \textbf{44.2} & \textbf{37.7} & \textbf{51.7} & \textbf{60.9} & \textbf{76.0} & \textbf{82.5}\\
\midrule
\abr{rcsls} & None & 14.6 & 17.1 & 5.0 & 18.3 & 19.2 & 43.6 & 50.5\\
& \abr{c+l} & 16.1 & 45.1 & 36.2 & 50.7 & 58.3 & 77.5 & 83.6\\
& \abr{in} & \textbf{56.3} & \textbf{48.6} & \textbf{38.0} & \textbf{52.4} & \textbf{60.5} & \textbf{78.1} & \textbf{83.9}\\
\bottomrule
\end{tabular}
\caption{Word translation accuracy aligning English embeddings to seven
  languages. %(more languages in Appendix~\ref{sec:result_all}).
  We combine three normalizations---no normalization (None), mean
  centering and length normalization (\abr{c+l}), and \name{}
  (\abr{in}) for five rounds---with three \abr{clwe}s: Procrustes, Procrustes
  with refinement~\citep{conneau-18}, and
  \abr{rcsls}~\citep{joulin-18}.
  Procrustes with \abr{c+l} is equivalent to \citet{artetxe-16}.
  The best result for each \abr{clwe} in each column \textbf{in
  bold}.  \name{} has the best accuracy of the three normalization techniques.}
\label{tab:result}
\end{table*}

\section{\name{}}

We now develop \name{}, which transforms monolingual word embeddings
to satisfy both length-invariance and center-invariance.
Specifically, we normalize word embeddings to simultaneously have
unit-length and zero-mean.
Formally, we produce embedding matrix $\vect{X}$ such that
\begin{equation}
  \| \vect{x}_i \|_2 = 1 \quad \text{for all $i$}, \label{eq:length-constraint}
\end{equation}
and
\begin{equation}
  \sum_{i=1}^n \vect{x}_i = \vect{0}. \label{eq:center-constraint}
\end{equation}

\name{} transforms the embeddings to make them satisfy both
constraints \textit{at the same time}.
Let $\vect{x}^{(0)}_i$ be the initial embedding for word $i$.
We assume that all word embeddings are non-zero.\footnote{For such vectors,
a small perturbation is an easy fix.}
For every word $i$, we iteratively transform each word vector
$\vect{x}_i$ by first making the vectors unit length,
\begin{equation}
  \vect{y}_i^{(k)} = \vect{x}_i^{(k-1)} / \| \vect{x}_i^{(k-1)} \|_2, \label{eq:renorm}
\end{equation}
  and then making them mean zero,
\begin{equation}
  \vect{x}_i^{(k)} = \vect{y}_i^{(k)}-\frac{1}{n} \sum_{i=1}^n \vect{y}_i^{(k)}.\label{eq:center}
\end{equation}

Equation~\eqref{eq:renorm} and \eqref{eq:center} project the embedding matrix
$\vect{X}$ to the set of embeddings that satisfy
Equation~\eqref{eq:length-constraint} and \eqref{eq:center-constraint}.
Therefore, our method is a form of alternating
projection~\citep{bauschke1996projection}, an algorithm to find a point in the
intersection of two closed sets by alternatively projecting onto one of the two
sets.
Alternating projection guarantees convergence in the intersection of
two convex sets at a linear rate~\cite{gubin1967method,
  bauschke1993convergence}.
Unfortunately, the unit-length constraint is \emph{non-convex}, ruling out the
classic convergence proof.
Nonetheless, we use recent results on alternating non-convex
projections~\cite{zhu2018convergence} to prove \name{}'s
convergence (details in Appendix~\ref{sec:proof}).

\begin{theorem}\label{thm:converge}
  If the embeddings are non-zero after each iteration; i.e., $\vect{x}_i^{(k)}
  \neq \vect{0}$ for all $i$ and $k$, then the sequence $\left\{ \vect{X}^{(k)}
  \right\}$ produced by \name{} is convergent.
\end{theorem}

All embeddings in our experiments satisfy the non-zero assumption; it is violated only when all words have the same embedding.
In degenerate cases, the algorithm might converge to a solution that does not
meet the two requirements.
Empirically, our method always satisfy both constraints.

\paragraph{Previous approach and differences.}
\citet{artetxe-16} also study he unit-length and zero-mean
constraints, but our work differs in two aspects.
First, they motivate the zero-mean condition based on the heuristic argument
that two randomly selected word types should not be semantically similar (or
dissimilar) in expectation.
While this statement is attractive at first blush, some word types have more
synonyms than others, so we argue that word types might not be evenly
distributed in the semantic space.
We instead show that zero-mean is helpful because it satisfies
center-invariance, a \textit{necessary condition} for orthogonal mappings.
Second, \citet{artetxe-16} attempt to enforce the two constraints by a single
round of dimension-wise mean centering and length normalization.
Unfortunately,
this often fails to meet the constraints \textit{at the same time}---length
normalization can change the mean, and mean centering can change
vector length.
In contrast, \name{} simultaneously meets both constraints and is
empirically better (Table~\ref{tab:result}) on dictionary induction.

\section{Dictionary Induction Experiments}\label{sec:exp}

On a dictionary induction benchmark, we combine \name{} with three
\abr{clwe} methods and show improvement in word translation accuracy across
languages.

\subsection{Dataset and Methods}

We train and evaluate \abr{clwe} on \abr{muse} dictionaries~\citep{conneau-18}
with default split.
We align English embeddings to thirty-nine target language embeddings,
pre-trained on Wikipedia with fastText~\cite{bojanowski-17}.  The alignment
matrices are trained from dictionaries of 5,000 source words.
We report top\nobreakdash-1 word translation accuracy for 1,500 source words,
using cross-domain similarity local scaling~\citep[\abr{csls}]{conneau-18}.
We experiment with the following \abr{clwe} methods.\footnote{We only report
accuracy for one run, because these \abr{clwe} methods are deterministic.}

\paragraph{Procrustes Analysis.}  Our first algorithm uses Procrustes
analysis~\cite{schonemann-66} to find the orthogonal transformation that
minimizes Equation~\ref{eq:obj}, the total distance between translation pairs.

\paragraph{Post-hoc Refinement.}  Orthogonal mappings can be improved
with refinement steps~\cite{artetxe-17,conneau-18}.
After learning an initial mapping $\vect{W}_0$ from the seed dictionary
$\mathcal{D}$, we build a synthetic dictionary $\mathcal{D}_1$ by translating
each word with $\vect{W}_0$.  We then use the new dictionary $\mathcal{D}_1$ to
learn a new mapping $\vect{W}_1$ and repeat the process.

\paragraph{Relaxed \abr{csls} Loss (\abr{rcsls}).}
\citet{joulin-18} optimize \abr{csls} scores between translation pairs instead
of Equation~\eqref{eq:obj}. 
\abr{rcsls} has state-of-the-art supervised word translation accuracies on
\abr{muse}~\citep{glavas-19}.
For the ease of optimization, \abr{rcsls} does not enforce the orthogonal
constraint.
Nevertheless, \name{} also improves its accuracy
(Table~\ref{tab:result}), showing it can help linear non-orthogonal
mappings too.

\vspace{.8em}
\subsection{Training Details}
\vspace{.4em}

We use the implementation from \abr{muse} for Procrustes analysis and
refinement~\citep{conneau-18}.  We use five refinement steps.
For \abr{rcsls}, we use the same hyperparameter selection strategy as
\citet{joulin-18}---we choose learning rate from $\{1, 10, 25, 50\}$ and number
of epochs from $\{10, 20\}$ by validation.
As recommended by~\citet{joulin-18}, we turn off the spectral constraint.
We use ten nearest neighbors when computing \abr{csls}.

\vspace{.8em}
\subsection{Translation Accuracy}
\vspace{.4em}

For each method, we compare three normalization strategies:
(1) no normalization,
(2) dimension-wise mean centering followed by length normalization~\citep{artetxe-16},
and (3) five rounds of \name{}.  
Table~\ref{tab:result} shows word translation accuracies on seven selected
target languages.  Results on other languages are in
Appendix~\ref{sec:result_all}.

As our theory predicts, \name{} increases translation accuracy for Procrustes
analysis (with and without refinement) across languages.
While centering and length-normalization also helps, the improvement is smaller,
confirming that one round of normalization is insufficient.
The largest margin is on English-Japanese, where \name{} increases test
accuracy by more than 40\%.
Figure~\ref{fig:example} shows an example of how \name{} makes the substructure
of an English-Japanese translation pair more similar. 

Surprisingly, normalization is even more important for \abr{rcsls}, a \abr{clwe}
method without orthogonal constraint.
\abr{rcsls} combined with \name{} has state-of-the-art accuracy,
but \abr{rcsls} is much worse than Procrustes analysis on unnormalized
embeddings,
suggesting that length-invariance and center-invariance are also helpful
for learning linear non-orthogonal mappings.

\begin{table}
\centering
\begin{tabular}{lrr}
  \toprule
  Dataset & Before & After\\
  \midrule
  \abr{ws-353} & {\bf 73.9} & 73.7\\ 
  \abr{mc} & 81.2 & {\bf 83.9}\\
  \abr{rg} & 79.7 & {\bf 80.0}\\
  \abr{yp-130} & 53.3 & {\bf 57.6}\\
  \bottomrule
\end{tabular}
\caption{Correlations before and after applying \name{} on four English word
  similarity benchmarks: \abr{ws-353}~\citep{finkelstein-02},
  \abr{mc}~\citep{miller-91}, \abr{rg}~\citep{rubenstein-65}, and
  \abr{yp-130}~\citep{yang-06}.
  The scores are similar, which shows that \name{} retains useful structures
  from the original embeddings.}
\label{tab:word-sim}
\end{table}

\newpage
\subsection{Monolingual Word Similarity}

Many trivial solutions satisfy both length-invariance and center-invariance;
e.g., we can map half of words to $\vect{e}$ and the rest to
$-\vect{e}$, where $\vect{e}$ is any unit-length vector.
A meaningful transformation should also preserve useful structure in the
original embeddings.
We confirm \name{} does not hurt scores on English word similarity
benchmarks~(Table~\ref{tab:word-sim}), showing that \name{} 
produces meaningful representations.

\section{Conclusion}

We identify two conditions that make cross-lingual orthogonal mapping easier:
length-invariance and center-invariance, and provide a simple algorithm that
transforms monolingual embeddings to satisfy both conditions.  Our method
improves word translation accuracy of different mapping-based \abr{clwe}
algorithms across languages.
In the future, we will investigate whether our method helps other downstream
tasks.

\section*{Acknowledgments}

We thank the anonymous reviewers for comments.
Boyd-Graber and Zhang are supported by DARPA award HR0011-15-C-0113 under
subcontract to Raytheon BBN Technologies.
Jegelka and Xu are supported by NSF CAREER award 1553284. Xu is also supported
by a Chevron-MIT Energy Fellowship. Kawarabayashi is supported by JST ERATO
JPMJER1201 and JSPS Kakenhi JP18H05291.
Any opinions, findings, conclusions, or recommendations expressed here are
those of the authors and do not necessarily reflect the view of the sponsors.

\clearpage

\bibliographystyle{style/acl_natbib}
\bibliography{bib/journal-full,bib/jbg,bib/mozhi,bib/keyulu}

\begin{thebibliography}{34}
\expandafter\ifx\csname natexlab\endcsname\relax\def\natexlab#1{#1}\fi

\bibitem[{Alvarez-Melis and Jaakkola(2018)}]{alvarez-18}
David Alvarez-Melis and Tommi~S. Jaakkola. 2018.
\newblock \href {https://www.aclweb.org/anthology/D18-1214} {Gromov-wasserstein
  alignment of word embedding spaces}.
\newblock In \emph{Proceedings of Empirical Methods in Natural Language
  Processing}.

\bibitem[{Alvarez-Melis et~al.(2019)Alvarez-Melis, Jegelka, and
  Jaakkola}]{alvarez-19}
David Alvarez-Melis, Stefanie Jegelka, and Tommi~S Jaakkola. 2019.
\newblock \href
  {http://proceedings.mlr.press/v89/alvarez-melis19a/alvarez-melis19a.pdf}
  {Towards optimal transport with global invariances}.
\newblock In \emph{Proceedings of Artificial Intelligence and Statistics}.

\bibitem[{Artetxe et~al.(2016)Artetxe, Labaka, and Agirre}]{artetxe-16}
Mikel Artetxe, Gorka Labaka, and Eneko Agirre. 2016.
\newblock \href {https://doi.org/10.18653/v1/D16-1250} {Learning principled
  bilingual mappings of word embeddings while preserving monolingual
  invariance}.
\newblock In \emph{Proceedings of Empirical Methods in Natural Language
  Processing}.

\bibitem[{Artetxe et~al.(2017)Artetxe, Labaka, and Agirre}]{artetxe-17}
Mikel Artetxe, Gorka Labaka, and Eneko Agirre. 2017.
\newblock \href {https://doi.org/10.18653/v1/P17-1042} {Learning bilingual word
  embeddings with (almost) no bilingual data}.
\newblock In \emph{Proceedings of the Association for Computational
  Linguistics}.

\bibitem[{Artetxe et~al.(2018)Artetxe, Labaka, and Agirre}]{artetxe-18b}
Mikel Artetxe, Gorka Labaka, and Eneko Agirre. 2018.
\newblock \href {https://www.aclweb.org/anthology/P18-1073} {A robust
  self-learning method for fully unsupervised cross-lingual mappings of word
  embeddings}.
\newblock In \emph{Proceedings of the Association for Computational
  Linguistics}.

\bibitem[{Bauschke and Borwein(1993)}]{bauschke1993convergence}
Heinz~H. Bauschke and Jonathan~M. Borwein. 1993.
\newblock On the convergence of von {N}eumann's alternating projection
  algorithm for two sets.
\newblock \emph{Set-Valued Analysis}, 1(2):185--212.

\bibitem[{Bauschke and Borwein(1996)}]{bauschke1996projection}
Heinz~H. Bauschke and Jonathan~M. Borwein. 1996.
\newblock \href {https://doi.org/10.1137/S0036144593251710} {On projection
  algorithms for solving convex feasibility problems}.
\newblock \emph{SIAM review}, 38(3):367--426.

\bibitem[{Bojanowski et~al.(2017)Bojanowski, Grave, Joulin, and
  Mikolov}]{bojanowski-17}
Piotr Bojanowski, Edouard Grave, Armand Joulin, and Tomas Mikolov. 2017.
\newblock \href {https://doi.org/10.1162/tacl_a_00051} {Enriching word vectors
  with subword information}.
\newblock \emph{Transactions of the Association for Computational Linguistics},
  5:135--146.

\bibitem[{Browder(1967)}]{browder-67}
Felix~E. Browder. 1967.
\newblock Convergence of approximants to fixed points of nonexpansive nonlinear
  mappings in {B}anach spaces.
\newblock \emph{Archive for Rational Mechanics and Analysis}, 24(1):82--90.

\bibitem[{Conneau et~al.(2018)Conneau, Lample, Ranzato, Denoyer, and
  J{\'e}gou}]{conneau-18}
Alexis Conneau, Guillaume Lample, Marc'Aurelio Ranzato, Ludovic Denoyer, and
  Herv{\'e} J{\'e}gou. 2018.
\newblock \href {https://openreview.net/forum?id=H196sainb} {Word translation
  without parallel data}.
\newblock In \emph{Proceedings of the International Conference on Learning
  Representations}.

\bibitem[{Doval et~al.(2018)Doval, Camacho-Collados, Espinosa-Anke, and
  Schockaert}]{doval-18}
Yerai Doval, Jose Camacho-Collados, Luis Espinosa-Anke, and Steven Schockaert.
  2018.
\newblock \href {https://www.aclweb.org/anthology/D18-1027} {Improving
  cross-lingual word embeddings by meeting in the middle}.
\newblock In \emph{Proceedings of Empirical Methods in Natural Language
  Processing}.

\bibitem[{Finkelstein et~al.(2002)Finkelstein, Gabrilovich, Matias, Rivlin,
  Solan, Wolfman, and Ruppin}]{finkelstein-02}
Lev Finkelstein, Evgeniy Gabrilovich, Yossi Matias, Ehud Rivlin, Zach Solan,
  Gadi Wolfman, and Eytan Ruppin. 2002.
\newblock \href {https://doi.org/10.1145/503104.503110} {Placing search in
  context: The concept revisited}.
\newblock \emph{ACM Transactions on information systems}, 20(1):116--131.

\bibitem[{Fujinuma et~al.(2019)Fujinuma, Boyd-Graber, and Paul}]{fujinuma-19}
Yoshinari Fujinuma, Jordan Boyd-Graber, and Michael~J. Paul. 2019.
\newblock A resource-free evaluation metric for cross-lingual word embeddings
  based on graph modularity.
\newblock In \emph{Proceedings of the Association for Computational
  Linguistics}.

\bibitem[{Glavas et~al.(2019)Glavas, Litschko, Ruder, and Vulic}]{glavas-19}
Goran Glavas, Robert Litschko, Sebastian Ruder, and Ivan Vulic. 2019.
\newblock How to (properly) evaluate cross-lingual word embeddings: On strong
  baselines, comparative analyses, and some misconceptions.
\newblock In \emph{Proceedings of the Association for Computational
  Linguistics}.

\bibitem[{Gubin et~al.(1967)Gubin, Polyak, and Raik}]{gubin1967method}
L.G. Gubin, B.T. Polyak, and E.V. Raik. 1967.
\newblock The method of projections for finding the common point of convex
  sets.
\newblock \emph{USSR Computational Mathematics and Mathematical Physics},
  7(6):1--24.

\bibitem[{Guo et~al.(2015)Guo, Che, Yarowsky, Wang, and Liu}]{guo-15}
Jiang Guo, Wanxiang Che, David Yarowsky, Haifeng Wang, and Ting Liu. 2015.
\newblock \href {https://doi.org/10.3115/v1/P15-1119} {Cross-lingual dependency
  parsing based on distributed representations}.
\newblock In \emph{Proceedings of the Association for Computational
  Linguistics}.

\bibitem[{Hoshen and Wolf(2018)}]{hoshen-18}
Yedid Hoshen and Lior Wolf. 2018.
\newblock \href {https://www.aclweb.org/anthology/D18-1043} {Non-adversarial
  unsupervised word translation}.
\newblock In \emph{Proceedings of Empirical Methods in Natural Language
  Processing}.

\bibitem[{Jawanpuria et~al.(2019)Jawanpuria, Balgovind, Kunchukuttan, and
  Mishra}]{jawanpuria-19}
Pratik Jawanpuria, Arjun Balgovind, Anoop Kunchukuttan, and Bamdev Mishra.
  2019.
\newblock \href {https://doi.org/10.1162/tacl_a_00257} {Learning multilingual
  word embeddings in latent metric space: a geometric approach}.
\newblock \emph{Transactions of the Association for Computational Linguistics},
  7:107--120.

\bibitem[{Joulin et~al.(2018)Joulin, Bojanowski, Mikolov, J{\'e}gou, and
  Grave}]{joulin-18}
Armand Joulin, Piotr Bojanowski, Tomas Mikolov, Herv{\'e} J{\'e}gou, and
  Edouard Grave. 2018.
\newblock \href {https://www.aclweb.org/anthology/D18-1330} {Loss in
  translation: Learning bilingual word mapping with a retrieval criterion}.
\newblock In \emph{Proceedings of Empirical Methods in Natural Language
  Processing}.

\bibitem[{Klementiev et~al.(2012)Klementiev, Titov, and
  Bhattarai}]{klementiev-12}
Alexandre Klementiev, Ivan Titov, and Binod Bhattarai. 2012.
\newblock \href {https://www.aclweb.org/anthology/C12-1089} {Inducing
  crosslingual distributed representations of words}.
\newblock \emph{Proceedings of International Conference on Computational
  Linguistics}.

\bibitem[{Mikolov et~al.(2013{\natexlab{a}})Mikolov, Le, and
  Sutskever}]{mikolov-13b}
Tomas Mikolov, Quoc~V. Le, and Ilya Sutskever. 2013{\natexlab{a}}.
\newblock \href {http://arxiv.org/abs/1309.4168} {Exploiting similarities among
  languages for machine translation}.
\newblock \emph{arXiv preprint arXiv:1309.4168}.

\bibitem[{Mikolov et~al.(2013{\natexlab{b}})Mikolov, Sutskever, Chen, Corrado,
  and Dean}]{mikolov-13-fixed}
Tomas Mikolov, Ilya Sutskever, Kai Chen, Gregory~S. Corrado, and Jeffrey Dean.
  2013{\natexlab{b}}.
\newblock \href
  {http://papers.nips.cc/paper/5021-distributed-representations-of-words-and-phrases-and-their-compositionality}
  {Distributed representations of words and phrases and their
  compositionality}.
\newblock In \emph{Proceedings of Advances in Neural Information Processing
  Systems}.

\bibitem[{Miller and Charles(1991)}]{miller-91}
George~A. Miller and Walter~G. Charles. 1991.
\newblock \href {https://doi.org/10.1080/01690969108406936} {Contextual
  correlates of semantic similarity}.
\newblock \emph{Language and Cognitive Processes}, 6(1):1--28.

\bibitem[{Pennington et~al.(2014)Pennington, Socher, and
  Manning}]{pennington-14}
Jeffrey Pennington, Richard Socher, and Christopher~D. Manning. 2014.
\newblock \href {https://doi.org/10.3115/v1/D14-1162} {{G}lo{V}e: Global
  vectors for word representation}.
\newblock In \emph{Proceedings of Empirical Methods in Natural Language
  Processing}.

\bibitem[{Rubenstein and Goodenough(1965)}]{rubenstein-65}
Herbert Rubenstein and John~B Goodenough. 1965.
\newblock \href {https://doi.org/10.1145/365628.365657} {Contextual correlates
  of synonymy}.
\newblock \emph{Communications of the ACM}, 8(10):627--633.

\bibitem[{Ruder et~al.(2018)Ruder, Cotterell, Kementchedjhieva, and
  S{\o}gaard}]{ruder-18}
Sebastian Ruder, Ryan Cotterell, Yova Kementchedjhieva, and Anders S{\o}gaard.
  2018.
\newblock \href {https://www.aclweb.org/anthology/D18-1042} {A discriminative
  latent-variable model for bilingual lexicon induction}.
\newblock In \emph{Proceedings of Empirical Methods in Natural Language
  Processing}.

\bibitem[{Ruder et~al.(2017)Ruder, Vulić, and S{\o}gaard}]{ruder-17}
Sebastian Ruder, Ivan Vulić, and Anders S{\o}gaard. 2017.
\newblock \href {http://arxiv.org/abs/1706.04902} {A survey of cross-lingual
  embedding models}.
\newblock \emph{arXiv preprint arXiv:1706.04902}.

\bibitem[{Sch{\"o}nemann(1966)}]{schonemann-66}
Peter~H. Sch{\"o}nemann. 1966.
\newblock A generalized solution of the orthogonal procrustes problem.
\newblock \emph{Psychometrika}, 31(1):1--10.

\bibitem[{Smith et~al.(2017)Smith, Turban, Hamblin, and Hammerla}]{smith-17}
Samuel~L. Smith, David H.~P. Turban, Steven Hamblin, and Nils~Y. Hammerla.
  2017.
\newblock \href {https://openreview.net/forum?id=r1Aab85gg} {Offline bilingual
  word vectors, orthogonal transformations and the inverted softmax}.
\newblock In \emph{Proceedings of the International Conference on Learning
  Representations}.

\bibitem[{S{\o}gaard et~al.(2018)S{\o}gaard, Ruder, and Vuli{\'c}}]{sogaard-18}
Anders S{\o}gaard, Sebastian Ruder, and Ivan Vuli{\'c}. 2018.
\newblock \href {https://www.aclweb.org/anthology/P18-1072} {On the limitations
  of unsupervised bilingual dictionary induction}.
\newblock In \emph{Proceedings of the Association for Computational
  Linguistics}.

\bibitem[{Xing et~al.(2015)Xing, Wang, Liu, and Lin}]{xing-15}
Chao Xing, Dong Wang, Chao Liu, and Yiye Lin. 2015.
\newblock \href {https://doi.org/10.3115/v1/N15-1104} {Normalized word
  embedding and orthogonal transform for bilingual word translation}.
\newblock In \emph{Conference of the North American Chapter of the Association
  for Computational Linguistics}.

\bibitem[{Yang and Powers(2006)}]{yang-06}
Dongqiang Yang and David~M. Powers. 2006.
\newblock Verb similarity on the taxonomy of wordnet.
\newblock In \emph{International {WordNet} Conference}.

\bibitem[{Zhang et~al.(2016)Zhang, Gaddy, Barzilay, and Jaakkola}]{zhang-16}
Yuan Zhang, David Gaddy, Regina Barzilay, and Tommi Jaakkola. 2016.
\newblock \href {https://doi.org/10.18653/v1/N16-1156} {Ten pairs to tag --
  multilingual {POS} tagging via coarse mapping between embeddings}.
\newblock In \emph{Conference of the North American Chapter of the Association
  for Computational Linguistics}.

\bibitem[{Zhu and Li(2018)}]{zhu2018convergence}
Zhihui Zhu and Xiao Li. 2018.
\newblock \href {http://arxiv.org/abs/1802.03889} {Convergence analysis of
  alternating nonconvex projections}.
\newblock \emph{arXiv preprint arXiv:1802.03889}.

\end{thebibliography}

\clearpage
\begin{appendix}
  \section{Proof for Theorem \ref{thm:converge}}\label{sec:proof}
Our convergence analysis is based on a recent result on alternating
non-convex projections. Theorem 1 in the work of ~\citet{zhu2018convergence}
states that the convergence of alternating projection holds even if the
constraint sets are non-convex, as long as the two constraint sets satisfy the following
assumption:

\begin{assumption}\label{asp:ap}
Let $ \mathbb{X}$ and $\mathbb{Y}$ be any two closed semi-algebraic sets, and let $\left\lbrace \left( \vect{x}_k, \vect{y}_k \right)  \right\rbrace $ be the sequence of iterates generated by the alternating projection method (e.g., \name{}). Assume the sequence $\left\lbrace \left( \vect{x}_k, \vect{y}_k \right)  \right\rbrace $ is bounded and the sets $ \mathbb{X}$ and $\mathbb{Y}$ obey the following properties: 
\begin{enumerate}
\item[(i)]  three-point property of $\mathbb{Y}$:  there exists a nonnegative function $\delta_{\alpha}: \mathbb{Y} \times \mathbb{Y} \rightarrow \mathbb{R}$ with $\alpha > 0$ such that for any $k \geq 1$, we have
\begin{equation*}
  \delta_{\alpha} \left( \vect{y}_k, \vect{y}_{k-1} \right) \geq \alpha  \|\vect{y}_k- \vect{y}_{k-1}\|_2
\end{equation*}
and 
\begin{equation*}
  \delta_{\alpha} \left(  \vect{y}_{k-1}, \vect{y}_k \right) + \| \vect{x}_k-\vect{y}_k \|_2^2 \\
  \leq  \| \vect{x}_k-\vect{y}_{k-1} \|_2^2, %& \\  \forall k \geq 1 &
\end{equation*}
\item[(ii)]  local contraction property of $ \mathbb{X}$:  there exist $\epsilon>0$ and $\beta>0$ such that when $\|\vect{y}_k-\vect{y}_{k-1}\|_2\leq \epsilon$, we have 
\begin{align*}
  \| \vect{x}_{k+1}-\vect{x}_k \|_2 & = \| \mathcal{P}_{ \mathbb{X}} (\vect{y}_k )- \mathcal{P}_{ \mathbb{X}} (\vect{y}_{k-1} )  \|_2 \\
  & \leq \beta \|  \vect{y}_k-\vect{y}_{k-1}  \|_2 
\end{align*}
where $\mathcal{P}_{ \mathbb{X}}$ is the projection onto ${ \mathbb{X}}$. 
\end{enumerate}
\end{assumption}

\citet{zhu2018convergence} only consider sets of vectors, but our constraint
are sets of matrices.  For ease of exposition, we treat every embedding
matrix $\vect{X} \in \mathbb{R}^{d \times n}$ as a vector by concatenating
the column vectors: $\vect{X} = \lbrack \vect{x}_1, \vect{x}_2, \cdots,
\vect{x}_n \rbrack$.  The $l^2$-norm of the concatenated vector $\| \vect{X}
\|_2$ is equivalent to the Frobenius norm of the original matrix $\| \vect{X}
\|_F$.

The two operations in \name{}, Equation~\eqref{eq:renorm} and \eqref{eq:center},
are projections onto two constraint sets, unit-length set $\mathbb{Y} =
\left\lbrace \vect{X} \in \mathbb{R}^{d \times n} : \forall i,
\|\vect{x}_i\|_2=1 \right\rbrace$ and zero-mean set $\mathbb{X} =
\left\lbrace \vect{X}\in\mathbb{R}^{d \times n} : \sum_{i=1}^n \vect{x}_i=\vect{0}
\right\rbrace$.  To prove convergence of \name{}, we show that $\mathbb{Y}$
satisfies the three-point property, and $\mathbb{X}$ satisfies the local
contraction property.

\paragraph{Three-point property of $\mathbb{Y}$.}
For any $\vect{Y}' \in \mathbb{Y}$ and $\vect{X} \in \mathbb{R}^{n\times d}$,
let $\vect{Y}$ be the projection of $\vect{X}$ onto the constraint set
$\mathbb{Y}$ with Equation~\eqref{eq:renorm}.
The columns of $\vect{Y}$ and $\vect{Y'}$ have the same length, so we have
\begin{align}
  &\| \vect{X}-\vect{Y}' \|_2^2-\| \vect{X}-\vect{Y} \|_2^2 \nonumber \\
  & \quad = \sum_{i=1}^n \| \vect{x}_i-\vect{y}'_i \|^2-\| \vect{x}_i-\vect{y}_i \|_2^2 \nonumber\\
  & \quad = \sum_{i=1}^n 2\vect{x}_i^\top\vect{y}_i-2\vect{x}_i^\top\vect{y}'_i.
  \label{eq:rewrite-1}
\end{align}
Since $\vect{Y}$ is the projection of $\vect{X}$ onto the unit-length
set with Equation~\eqref{eq:renorm}; i.e., $\vect{y}_i = \vect{x}_i / \|\vect{x}_i\|_2$,
we can rewrite Equation~\eqref{eq:rewrite-1}.
\begin{align}
  &\| \vect{X}-\vect{Y}' \|_2^2-\| \vect{X}-\vect{Y} \|_2^2 \nonumber\\
  &\quad = \sum_{i=1}^n \|\vect{x}_i\|_2 (2\vect{y}_i^\top\vect{y}_i-2\vect{y}_i^\top\vect{y}'_i). \label{eq:rewrite-2}
\end{align}
All columns of $\vect{Y}$ and $\vect{Y'}$ are unit-length.  Therefore, we can
further rewrite Equation~\eqref{eq:rewrite-2}.
\begin{align*}
  &\| \vect{X}-\vect{Y}' \|_2^2-\| \vect{X}-\vect{Y} \|_2^2\\
  & \quad = \sum_{i=1}^n \| \vect{x}_i \|_2 (2- 2 \vect{y}_i^\top \vect{y}'_i)\\
  & \quad = \sum_{i=1}^n \| \vect{x}_i \|_2 \| \vect{y}_i-\vect{y}'_i \|_2^2.
\end{align*}
Let $l = \min_{i} \left\lbrace \|\vect{x}_i\|_2 \right\rbrace$ be the minimum
length of the columns in $\vect{X}$.  We have the following inequality:
\begin{align*}
  &\| \vect{X}-\vect{Y}' \|_2^2-\| \vect{X}-\vect{Y} \|_2^2\\
  & \quad \geq \sum_{i=1}^n l \| \vect{y}_i-\vect{y}'_i \|_2^2\\
  & \quad = l ||\vect{Y}-\vect{Y}'\|_2^2.
\end{align*}
From our non-zero assumption, the minimum column length $l$ is always positive.
Let $l_k$ be the minimum column length of the embedding matrix
$\vect{X}^{(k)}$ after the $k$-th iteration.  It follows that $\mathbb{Y}$
satisfies the three-point property with $\alpha = \min_{k} \left\lbrace l_k
\right\rbrace$ and $\delta_\alpha (\vect{Y}, \vect{Y}') = \alpha \| \vect{Y}
-\vect{Y}'\|_2^2$.

\paragraph{Local contraction property of $\mathbb{X}$.}
The zero-mean constraint set $\mathbb{X}$ is convex and closed: if two
matrices $\vect{X}$ and $\vect{Y}$ both have zero-mean, their linear
interpolation $\lambda \vect{X} + (1 - \lambda) \vect{Y}$ must also have
zero-mean for any $0 < \lambda < 1$.
Projections onto convex sets in a Hilbert space are
contractive~\cite{browder-67}, and therefore $\mathbb{X}$ satisfies the local
contraction property with any positive $\epsilon$ and $\beta = 1$.

In summary, the two constraint sets that \name{} projects onto satisfy
Assumption~\ref{asp:ap}.  Therefore, \name{} converges following the analysis 
of~\citet{zhu2018convergence}.

\section{Results on All Languages}\label{sec:result_all}

Table~\ref{tab:result_all} shows word translation accuracies on all target
languages.  \name{} improves accuracy on all languages.

\begin{table*}[p]
  \centering
  \setlength{\tabcolsep}{0.1cm}
  \begin{tabular}{lRRRRRRRRR}
    \toprule
    & \multicolumn{3}{c}{Procrustes} & \multicolumn{3}{c}{Procrustes + refine} & \multicolumn{3}{c}{\abr{rcsls}}\\
    \cmidrule(lr){2-4}\cmidrule(lr){5-7}\cmidrule(lr){8-10}
    Target & None & \abr{c+l} & \abr{in} & None & \abr{c+l} & \abr{in} & None & \abr{c+l} & \abr{in}\\
    \midrule
    \abr{af} & 26.3 & 28.3 & \bf{29.7} & 27.7 & 28.7 & \bf{30.4} & 9.3 & 28.6 & \bf{29.3}\\
    \abr{ar} & 36.5 & 37.1 & \bf{37.9} & 36.5 & 37.1 & \bf{37.9} & 18.4 & 40.5 & \bf{41.5}\\
    \abr{bs} & 22.3 & 23.5 & \bf{24.4} & 23.3 & 23.9 & \bf{26.6} & 5.4 & 25.5 & \bf{26.6}\\
    \abr{ca} & 65.9 & 67.6 & \bf{68.9} & 66.5 & 67.6 & \bf{68.9} & 43.0 & 68.9 & \bf{69.5}\\
    \abr{cs} & 54.0 & 54.7 & \bf{55.3} & 54.0 & 54.7 & \bf{55.7} & 29.9 & 57.8 & \bf{58.2}\\
    \abr{da} & 54.0 & 54.9 & \bf{58.4} & 56.8 & 59.3 & \bf{60.9} & 19.2 & 58.3 & \bf{60.5}\\
    \abr{de} & 73.5 & 74.6 & \bf{75.5} & 74.3 & 75.2 & \bf{76.0} & 43.6 & 77.5 & \bf{78.1}\\
    \abr{el} & 44.0 & 44.9 & \bf{47.5} & 44.6 & 45.9 & \bf{47.9} & 14.0 & 47.1 & \bf{48.5}\\
    \abr{es} & 81.4 & 81.3 & \bf{81.5} & 81.9 & 82.1 & \bf{82.5} & 50.5 & 83.6 & \bf{83.9}\\
    \abr{et} & 31.9 & 34.5 & \bf{36.1} & 31.9 & 35.3 & \bf{36.4} & 8.1 & 37.3 & \bf{39.4}\\
    \abr{fa} & 33.1 & 33.7 & \bf{37.3} & 33.1 & 34.1 & \bf{37.3} & 5.9 & 37.5 & \bf{38.3}\\
    \abr{fi} & 47.6 & 48.5 & \bf{50.9} & 47.6 & 50.1 & \bf{51.1} & 20.9 & 52.3 & \bf{53.3}\\
    \abr{fr} & 81.1 & 81.3 & \bf{81.7} & 82.1 & 82.7 & \bf{82.4} & 53.1 & 83.9 & \bf{83.9}\\
    \abr{he} & 40.2 & 43.1 & \bf{43.7} & 40.2 & 43.1 & \bf{43.7} & 13.1 & 49.7 & \bf{50.1}\\
    \abr{hi} & 33.3 & 34.0 & \bf{36.7} & 33.6 & 34.9 & \bf{37.7} & 5.0 & 36.2 & \bf{38.0}\\
    \abr{hr} & 37.0 & 37.8 & \bf{40.2} & 37.6 & 37.8 & \bf{40.2} & 14.5 & 41.1 & \bf{42.6}\\
    \abr{hu} & 51.8 & 54.1 & \bf{55.5} & 53.3 & 54.1 & \bf{56.1} & 11.7 & 57.3 & \bf{58.2}\\
    \abr{id} & 65.6 & 65.7 & \bf{67.9} & 67.7 & 68.4 & \bf{70.3} & 24.8 & 68.9 & \bf{70.0}\\
    \abr{it} & 76.2 & 76.6 & \bf{76.6} & 77.5 & 78.1 & \bf{78.1} & 48.4 & 78.8 & \bf{79.1}\\
    \abr{ja} & 1.7 & 13.1 & \bf{44.3} & 1.7 & 13.1 & \bf{44.3} & 14.6 & 16.1 & \bf{56.3}\\
    \abr{ko} & 31.5 & 32.1 & \bf{33.9} & 31.5 & 32.1 & \bf{33.9} & 6.4 & 37.5 & \bf{37.5}\\
    \abr{lt} & 22.5 & 22.8 & \bf{23.2} & 22.5 & 22.8 & \bf{23.3} & 7.6 & 23.3 & \bf{23.5}\\
    \abr{lv} & 23.6 & 24.9 & \bf{26.1} & 23.6 & 24.9 & \bf{26.1} & 10.1 & 28.3 & \bf{28.7}\\
    \abr{ms} & 44.0 & 45.4 & \bf{48.9} & 46.5 & 48.3 & \bf{51.1} & 19.9 & 49.1 & \bf{50.2}\\
    \abr{nl} & 72.8 & 73.7 & \bf{74.1} & 73.8 & 75.1 & \bf{75.8} & 46.7 & 75.6 & \bf{75.8}\\
    \abr{pl} & 58.2 & \bf{60.2} & 60.1 & 58.5 & 60.2 & \bf{60.4} & 39.4 & 62.4 & \bf{62.5}\\
    \abr{pt} & 79.5 & 79.7 & \bf{79.9} & 79.9 & 81.0 & \bf{81.2} & 63.1 & 81.1 & \bf{81.7}\\
    \abr{ro} & 58.1 & 60.5 & \bf{61.8} & 59.9 & 60.5 & \bf{62.5} & 27.1 & 61.9 & \bf{63.3}\\
    \abr{ru} & 51.7 & 52.1 & \bf{52.1} & 51.7 & 52.1 & \bf{52.1} & 26.6 & 57.1 & \bf{57.9}\\
    \abr{sk} & 38.0 & 39.3 & \bf{40.4} & 38.0 & 39.3 & \bf{41.7} & 13.3 & 41.5 & \bf{42.3}\\
    \abr{sl} & 32.5 & 34.3 & \bf{36.7} & 32.5 & 34.4 & \bf{36.7} & 12.3 & 36.0 & \bf{37.9}\\
    \abr{sq} & 23.5 & 25.1 & \bf{27.3} & 23.5 & 25.1 & \bf{27.3} & 4.4 & 26.5 & \bf{27.3}\\
    \abr{sv} & 58.7 & 59.6 & \bf{60.7} & 60.9 & 61.2 & \bf{62.6} & 35.6 & 63.8 & \bf{63.9}\\
    \abr{ta} & 15.1 & 15.5 & \bf{16.8} & 15.1 & 15.5 & \bf{17.7} & 6.7 & 16.3 & \bf{17.1}\\
    \abr{th} & 22.5 & \bf{23.3} & 22.9 & 22.5 & \bf{23.3} & 22.9 & 9.4 & 23.7 & \bf{23.9}\\
    \abr{tr} & 44.9 & 46.5 & \bf{48.7} & 46.3 & 48.7 & \bf{51.7} & 18.3 & 50.7 & \bf{52.4}\\
    \abr{uk} & 34.8 & 35.9 & \bf{36.3} & 35.5 & 35.9 & \bf{36.5} & 18.8 & 40.7 & \bf{40.8}\\
    \abr{vi} & 41.3 & 42.1 & \bf{43.7} & 42.1 & 42.7 & \bf{44.2} & 14.2 & 43.3 & \bf{43.9}\\
    \abr{zh} & 32.5 & 42.3 & \bf{44.2} & 32.5 & 42.3 & \bf{44.2} & 17.1 & 45.1 & \bf{48.6}\\
    \midrule
    Average & 44.7 & 46.3 & \bf{48.4} & 45.3 & 47.0 & \bf{49.1} & 21.8 & 49.0 & \bf{50.9}\\
    \bottomrule
  \end{tabular}
  \caption{Word translation accuracy aligning English embeddings to thirty-nine
  languages.
  We combine three normalizations---no normalization (None), mean
  centering and length normalization (\abr{c+l}), and \name{}
  (\abr{in}) for five rounds---with three \abr{clwe}s: Procrustes, Procrustes
  with refinement~\citep{conneau-18}, and
  \abr{rcsls}~\citep{joulin-18}.
  Procrustes with \abr{c+l} is equivalent to \citet{artetxe-16}.
  The best result for each \abr{clwe} in each column \textbf{in
  bold}.  \name{} has the best accuracy of the three normalization techniques.}
  \label{tab:result_all}
\end{table*}

\end{appendix}

\end{document}